\documentclass[conference]{IEEEtran}
\usepackage{cite}
\usepackage{amsmath,amssymb,amsfonts}
\usepackage{algorithmic}
\usepackage{graphicx}
\usepackage{textcomp}
\usepackage{xcolor}
\graphicspath{{./images/}} 
\usepackage{graphicx}
\usepackage{multirow}
\usepackage{xspace}
\usepackage{subcaption} 
\usepackage{caption}
\usepackage{textcomp} 
\usepackage{hyperref}  
\hypersetup{
    colorlinks=true,        
    linkcolor=black,         
    urlcolor=black,           
    citecolor=black,        
    pdfborder={0 0 0},      
}

\usepackage{array} 
\usepackage{multirow} 
\usepackage{booktabs} 
\usepackage{siunitx}  
\usepackage{amsmath}  
\usepackage{url} 

\def\BibTeX{{\rm B\kern-.05em{\sc i\kern-.025em b}\kern-.08em
    T\kern-.1667em\lower.7ex\hbox{E}\kern-.125emX}}
\begin{document}

\newcommand{\hm}[1]{\textcolor{teal}{Haomin: [#1]}}

\newcommand{\kencoder}{\textbf{kernel encoder}\xspace}
\newcommand{\kdecoder}{\textbf{kernel decoder}\xspace}
\newcommand{\obsencoder}{\textbf{observation and activation map encoder}\xspace}

\title{Quasiparticle Interference Kernel Extraction with Variational Autoencoders via Latent Alignment
}


\IEEEoverridecommandlockouts  

\author{
\IEEEauthorblockN{
Yingshuai Ji\textsuperscript{*}\textsuperscript{\dag}
}
\IEEEauthorblockA{\textit{Department of Computer Science \& } \\\textit{Engineering} \\
\textit{University of Notre Dame}\\
South Bend, IN, USA \\
ysji0505@gmail.com}
\and
\IEEEauthorblockN{
Haomin Zhuang\textsuperscript{*}
}
\IEEEauthorblockA{\textit{Department of Computer Science \& } \\\textit{Engineering}  \\
\textit{University of Notre Dame}\\
South Bend, IN, USA \\
hzhuang2@nd.edu}
\and
\IEEEauthorblockN{
Matthew Toole}
\IEEEauthorblockA{\textit{Department of Physics \& Astronomy} \\
\textit{University of Notre Dame}\\
South Bend, IN, USA \\
mtoole2@nd.edu}
\and
\IEEEauthorblockN{
James McKenzie}
\IEEEauthorblockA{\textit{Department of Physics \& Astronomy} \\
\textit{University of Notre Dame}\\
South Bend, IN, USA \\
jmckenz2@nd.edu}
\and
\IEEEauthorblockN{
Xiaolong Liu\textsuperscript{\ddag}
}
\IEEEauthorblockA{\textit{Department of Physics \& Astronomy} \\
\textit{University of Notre Dame}\\
South Bend, IN, USA \\
xliu33@nd.edu}
\and
\IEEEauthorblockN{
Xiangliang Zhang\textsuperscript{\ddag}
}
\IEEEauthorblockA{\textit{Department of Computer Science \& } \\\textit{Engineering} \\
\textit{University of Notre Dame}\\
South Bend, IN, USA \\
xzhang33@nd.edu}

\thanks{\textsuperscript{*}These authors contributed equally to this work.}
\thanks{\textsuperscript{\dag}Work done via internship at University of Notre Dame.}
\thanks{\textsuperscript{\ddag}Corresponding authors.}
}

\maketitle
\begin{abstract}
Quasiparticle interference (QPI) imaging is a powerful tool for probing electronic structures in quantum materials, but extracting the single-scatterer QPI pattern (i.e., the \emph{kernel}) from a multi-scatterer image remains a fundamentally ill-posed inverse problem, because many different kernels can combine to produce almost the same observed image, and noise or overlaps further obscure the true signal. Existing solutions to this extraction problem rely on manually zooming into small local regions with isolated single-scatterers. This is infeasible for real cases where scattering conditions are too complex. 
In this work, we propose the first AI-based framework for QPI kernel extraction, which models the space of physically valid kernels and uses this knowledge to guide the inverse mapping. We introduce a two-step learning strategy that decouples kernel representation learning from observation-to-kernel inference. In the first step, we train a variational autoencoder to learn a compact latent space of scattering kernels. In the second step, we align the latent representation of QPI observations with those of the pre-learned kernels using a dedicated encoder. This design enables the model to infer kernels robustly under complex, entangled scattering conditions. We construct a diverse and physically realistic QPI dataset comprising 100 unique kernels and evaluate our method against a direct one-step baseline. Experimental results demonstrate that our approach achieves significantly higher extraction accuracy, improved generalization to unseen kernels.  To further validate its effectiveness, we also apply the method to real QPI data from Ag and FeSe samples, where it reliably extracts meaningful kernels under complex scattering conditions. Codes and data are available at \href{https://github.com/QPI-Dataset/QPI-Dataset-100kernels}{\textcolor{blue}{\textit{https://github.com/QPI-Dataset/QPI-Dataset-100kernels}}}.
\end{abstract}


\section{INTRODUCTION}

Understanding the behavior of electrons in quantum materials is a central challenge in modern condensed matter physics, with far-reaching implications for the development of superconductors, and other next-generation electronic systems~\cite{Liu_2024,keimer2017high,doi:10.1021/acs.chemrev.0c01322,Ball_2017}. One powerful experimental technique used to probe the electronic properties of such materials is \textit{quasiparticle interference} (QPI) imaging 
via scanning tunneling microscopy (STM), which captures quantum interference patterns that arise from electrons elastically scattering off impurities in a material (usually characterized as atomic defects) \cite{crommie1993imaging,Arguello_2015}. 
The core of QPI analysis is to identify the interference pattern, or \textbf{kernel}, associated with a \textbf{single scatterer}, as this pattern can be used to extract momentum-space electronic structure using a Fourier transform\cite{Sharma_2021}. However, experimentally, it is challenging to find such single isolated defects in a sufficiently large defect-free region, due to the ubiquitousness of surface defects\cite{doi:10.1126/science.1187399,Allan_2013}. Scientists thus have to analyze \textbf{QPI images with multiple defects}, where each defect acts as a scatterer that contributes to the overall interference pattern. 

\begin{figure}[t]
  \centering
  \includegraphics[width=0.72\linewidth]{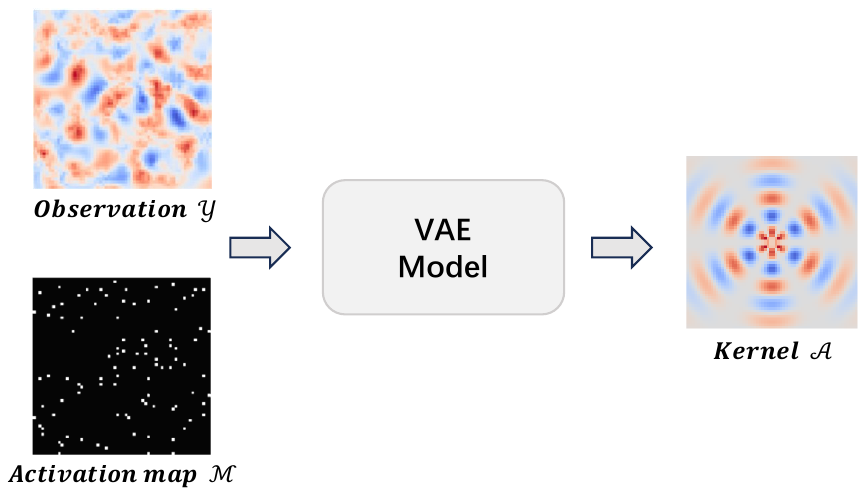} 
  \caption{Illustration of the QPI Kernel Extraction Problem} 
  \label{fig:example}
    \vspace{-0.2in}
\end{figure}

However, because of their random distribution, QPI phase incoherence among scatterers greatly reduces the overall signal-to-noise ratio (S/N) of a multi-scatterer QPI pattern \cite{cheung2020dictionary}.  A useful physical analogy is to imagine a random distribution of many identical stones being dropped into a pool of water, and using the resulting superposition of ripples to deduce the ripple pattern from a single stone.   
In our QPI kernel extraction problem,  as illustrated in Fig.~\ref{fig:example}, the task is to \textbf{deconvolute and infer the kernel $\mathcal{A}$ based on a multi-scatterer QPI image (observation) $\mathcal{Y}$ and the known defect locations (activation map) $\mathcal{M}$}.  

Existing solutions to this QPI kernel extraction problem are either zoom into each isolated defect in a small defect-free region \cite{sharma2021multi}, or try to extract a dispersion relation in a field of view with many defects directly \cite{hoffman2002imaging}. 
When images are taken in a small field of view, momentum space resolution is reduced. Therefore, for images with many defects generating overlapping QPI patterns, the compromised S/N prevents the extraction of momentum space electronic structure with highest precision\cite{Zhang_2020,hsu2017unsupervised}. This motivates our strategy of designing \textbf{machine learning approaches to analyze images} of large regions with \textbf{multiple defects and extract the kernel}.  

We propose to leverage the power of convolutional neural networks (CNNs)~\cite{lecun2002gradient} to learn spatial patterns in the observed QPI signal and the expressive latent representations of a variational autoencoder (VAE) \cite{kingma2014auto,kingma2019introduction} to infer the underlying scattering kernel. 
A straightforward solution is to employ an encoder-decoder that jointly takes  $\mathcal{Y}$ and  $\mathcal{M}$ as input and directly decodes the scattering kernel $\mathcal{A}$ (e.g., using a single VAE in the solution shown by Fig.~\ref{fig:example}).
However, this direct decoding approach may suffer from poor performance due to the difficulty in learning a stable mapping from observations with overlapping kernels to a single kernel in a highly entangled space, as we demonstrate later in our evaluation results.

To address these challenges, we propose a \textbf{two-step solution that decomposes the learning task into two more manageable subproblems}\cite{ilse2019divadomaininvariantvariational,wang2021twin,zhu2021learningaudio}, as shown in Fig. \ref{fig:framework} (Top). In the \textbf{first} step, we train a variational autoencoder where the scattering kernel $\mathcal{A}$ is both the input and the target output. This step learns a \textbf{compact, semantically meaningful embedding space for kernels, $\textbf{h}_A$}.  
In the \textbf{second} step, we aim to \textbf{bridge the observation and kernel domains in their representation space}. Specifically, we train a separate encoder that maps the observational data $(\mathcal{Y}, \mathcal{M})$ to a latent embedding $\textbf{h}_y$, which is then aligned with the latent code ($\textbf{h}_A$) of the corresponding kernel produced by the step1-trained (now frozen) encoder. By enforcing this alignment, the model learns to infer kernel representations from observations, enabling robust kernel extraction via the shared decoder (Fig. \ref{fig:framework} Bottom).

Considering that real QPI experiments are costly and lack accessible ground-truth kernels, we initiate the study by constructing a simulated QPI dataset to facilitate broader AI research in this domain and to enable systematic evaluation of kernel extraction methods. Our simulated datasets cover a wide range of scattering environments and electronic structures, involving 100 distinct scattering kernels. This dataset is designed to reflect the diversity and complexity encountered in real QPI measurements (though different from the real world), including variations in impurity configurations and band structures. To reduce the gap between simulation and real-world QPI experiment, we then further introduce 
controlled noise to the simulated data, enabling the model to handle the noise characteristics present in real measurements. 

To evaluate the effectiveness of the proposed model, we conducted kernel inference on both simulated datasets and real QPI samples of Ag and FeSe. 
Experimental results indicate that our proposed method significantly outperforms the direct one-step baseline (e.g., a single VAE in Fig.~\ref{fig:example}) in terms of extraction accuracy and generalization to unseen kernels. Importantly, validation on real QPI data shows qualitative consistency with our simulated experiments, further supporting both the robustness of our approach and the effectiveness of the constructed dataset in capturing the essential characteristics of real QPI measurements. These results highlight the fundamental limitation of direct one-step solutions: they attempt to solve the ill-posed inverse problem without sufficient structural priors. Our two-step approach, in contrast, explicitly disentangles kernel representation learning from observation-to-kernel inference, which makes the inversion far more stable.
 
This can be better understood through analogy: the one-step solution is akin to recovering the ripple pattern of a single stone from a snapshot of overlapping waves,  without first understanding the shape of individual ripples. Small shifts in stone positions can drastically alter the observed interference, even if the ripple shape remains unchanged. Similarly, in QPI, the observed signal reflects entangled scattering effects, making direct inference unstable. Our two-step method addresses this by first learning the structure of individual kernels, then aligning observations within the same representation space to enable robust and generalizable extraction.

Overall, the main contributions of this paper are as follows:

\begin{itemize}
    \item  We formulate QPI kernel extraction as a challenging inverse problem and propose the first machine learning-based solution to this task. Our principled two-step framework decouples kernel representation learning from observation-to-kernel inference, marking a step toward data-driven approaches for inverse problems in quantum materials research. More broadly, this work illustrates how modern AI techniques can unlock new opportunities in scientific domains traditionally limited by small, noisy, and ill-posed datasets.
    \item We construct a diverse QPI dataset comprising 100 distinct scattering kernels across a variety of electronic structures and impurity configurations,  with 500 observations each for a total of 50,000 samples. To better approximate experimental conditions, we further enhance the dataset through systematic noise augmentation. In addition, we collect a small set of real QPI measurements on Ag and FeSe as a validation set to demonstrate the dataset’s effectiveness and similarity to experimental data. We release the dataset in public
 to facilitate future research and benchmarking in machine learning for quantum materials.
    
    \item We demonstrate that our two-step method consistently outperforms the direct one-step baseline up to 18\% of Root Mean Square Error (RMSE), presenting the first state-of-the-art AI-based solution to the QPI kernel extraction problem. 
\end{itemize}

\section{RELATED WORKS}
\subsection{Quasiparticle Interference (QPI) Imaging}
Quasiparticle interference (QPI) imaging is a powerful STM-based technique that captures spatial oscillations in the local density of states induced by quasiparticle scattering from defects and impurities. By Fourier transforming these patterns, QPI provides momentum-resolved insights into electronic band dispersions, Fermi surface topology, and scattering processes, thereby bridging real-space imaging with reciprocal-space spectroscopy. This capability has made QPI indispensable for studying emergent phenomena in quantum materials, including high-$T_c$ superconductivity, topological surface states, and correlated electron behavior. Early QPI studies focused on quasi-2D metals such as Cu(111) and Au(111), and later extended to cuprates, iron-based superconductors, and topological insulators. More recently, theoretical advances have expanded QPI analysis to three-dimensional materials using Green’s function methods to model subsurface scattering~\cite{rhodes2023nature}. Computational efforts such as wave-packet simulations have been employed to reproduce intervalley and intravalley scattering features in graphene~\cite{vancso2021wave}. On the experimental side, innovations like projective QPI (PQPI) allow one-dimensional high-resolution scans centered on single defects to enhance energy-dispersion mapping~\cite{zhang2020projective}, while adaptive sparse sampling has been proposed to accelerate STM-based QPI acquisition without loss of information~\cite{oppliger2022adaptive}. In parallel, simulation frameworks such as calcQPI leverage tight-binding Green’s functions to efficiently model QPI spectra from realistic Hamiltonians~\cite{wahl2025calcqpi}. Despite these advances, extracting the intrinsic single-scatterer QPI kernel from multi-scatterer patterns remains a fundamentally ill-posed inverse problem.

\subsection{Data-driven Solutions for Scientific Inverse Problems}
Artificial intelligence has recently emerged as a powerful paradigm for addressing inverse problems across scientific domains, where the goal is to reconstruct hidden physical quantities from indirect or incomplete measurements. Traditional inverse problem approaches, such as regularized optimization, compressed sensing, and Bayesian inference, often require strong prior assumptions and are computationally intensive, especially in high-dimensional or ill-posed settings. By contrast, AI-based methods leverage data-driven priors and flexible neural architectures to learn direct mappings between observations and underlying parameters, enabling faster and more robust inference. In medical imaging, deep neural networks have been employed for accelerated magnetic resonance imaging (MRI) reconstruction~\cite{zhu2018image} and low-dose computed tomography (CT) denoising~\cite{macmahon2017guidelines}. In physics and materials science, generative models and variational autoencoders have been used to infer electronic band structures and reconstruct scattering potentials from spectroscopic data~\cite{carleo2019machine}, as well as to design material parameters from desired spectral responses \cite{ahmed2021deterministic,ahmed2023machine,ahmed2023generative,ahmed2025machine}. Similar advances have been demonstrated in geophysics, where deep learning accelerates seismic inversion by approximating wave-equation solvers~\cite{das2019convolutional}. More recently, physics-informed neural networks (PINNs) have integrated physical laws into learning architectures to enforce consistency with governing equations, providing a general framework for solving PDE-constrained inverse problems~\cite{raissi2019physics}. Collectively, these developments highlight the transformative role of data-driven approaches in advancing inverse problem solving.

\subsection{Variational Autoencoders}
Generative modeling has seen rapid progress with several mainstream frameworks, including Generative Adversarial Networks (GANs), Normalizing Flows, Diffusion Models, and Variational Autoencoders (VAEs). Among them, VAEs provide a lightweight and probabilistic formulation that enables stable training, compact latent representations, and interpretable feature spaces. VAEs have thus emerged as one of the most influential generative modeling frameworks since their introduction by \cite{kingma2014auto} and \cite{rezende2014stochastic}, offering a probabilistic approach to representation learning and efficient latent space inference. VAEs have been widely applied to image synthesis~\cite{yan2016attribute2image,larsen2016autoencoding}, text generation~\cite{miao2016neural}, and speech modeling~\cite{hsu2017unsupervised,chorowski2019unsupervised}, highlighting their ability to learn compact latent structures of high-dimensional data. Beyond these classical domains, VAEs have gained increasing attention in scientific applications. In the physical sciences, VAEs have been employed to analyze spectroscopic and scattering data, enabling compact latent representations of electronic band structures and scattering kernels that are difficult to extract through traditional inverse methods~\cite{carleo2019machine,ahmed2021deterministic,ahmed2023machine}. In chemistry and molecular design, \cite{gomez2018automatic} pioneered the use of VAEs to generate novel molecular structures by navigating continuous latent spaces, while subsequent studies extended these approaches to property optimization and retrosynthesis~\cite{polykovskiy2020molecular}. Similarly, VAEs have been adapted to protein science, where they facilitate structure generation and functional design \cite{riesselman2017deep}. These works collectively demonstrate that VAEs are not only powerful in conventional generative modeling, but also play an increasingly critical role in accelerating scientific discovery by providing interpretable latent spaces and enabling data-driven exploration of complex physical and chemical systems.

\begin{figure}[t]
  \centering
  \includegraphics[width=0.95\linewidth]{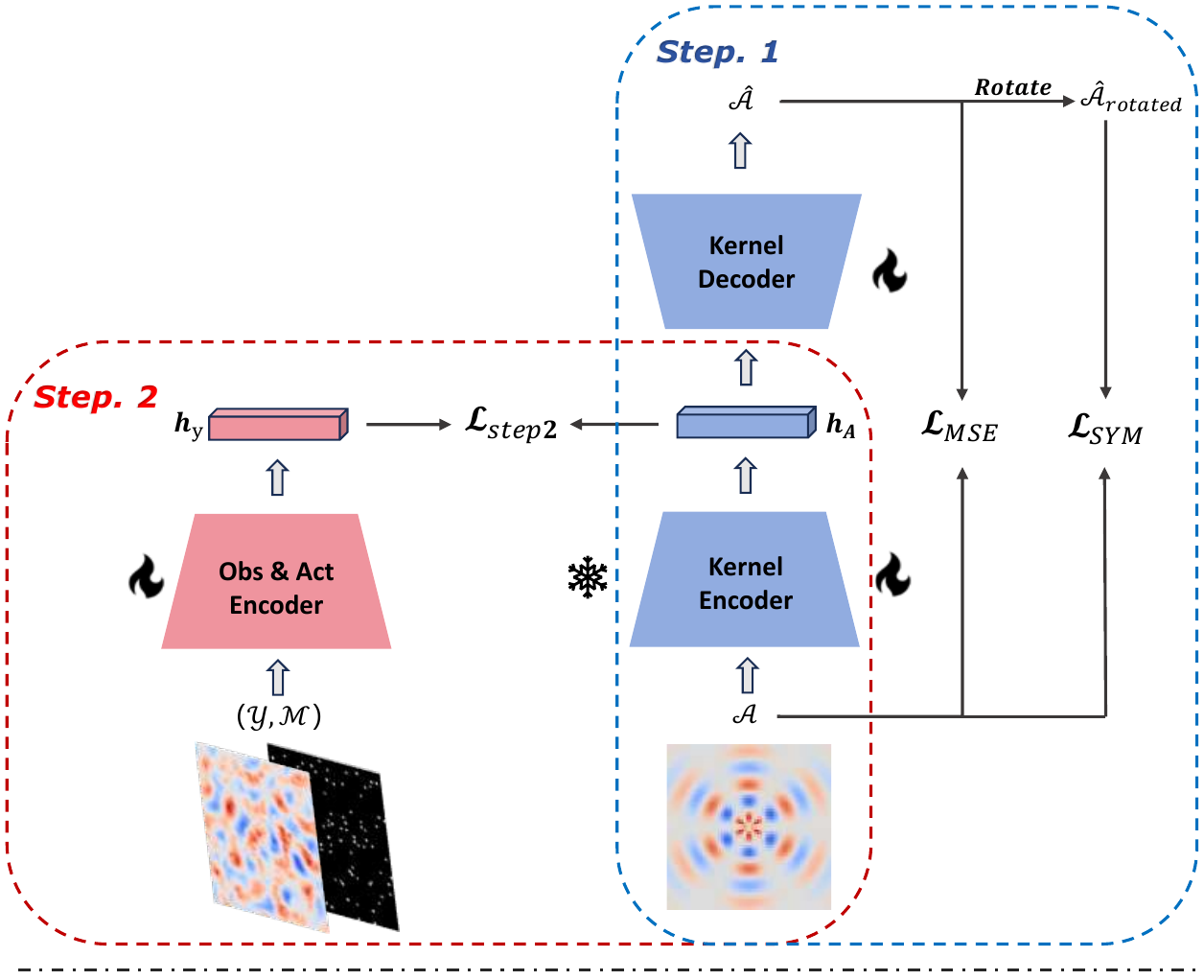}

  \includegraphics[width=0.7\linewidth]{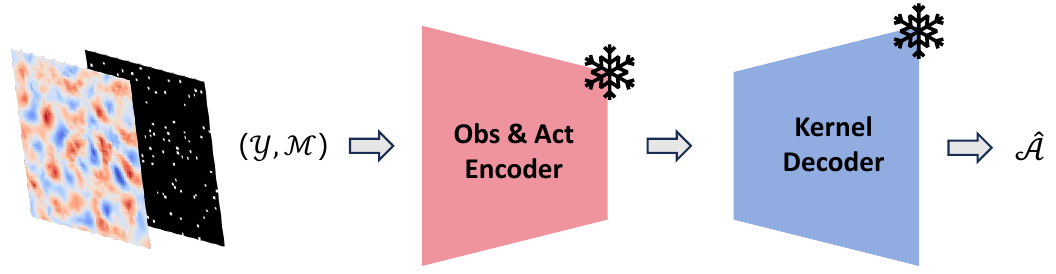}
  \caption{\underline{Top}: The proposed two-step training framework.
  \textbf{Step 1} (blue box): training a VAE for kernel~$\mathcal{A}$ reconstruction  using the losses $\mathcal{L}_{\text{MSE}}$ and $\mathcal{L}_{\text{SYM}}$. 
     \textbf{Step 2} (red box): training the observation–activation encoder via aligning $\mathbf{h}_A$ and $\mathbf{h}_y$ (the kernel encoder is frozen). 
\underline{Bottom}: The inference process.}

  \label{fig:framework}
  \vspace{-0.2in}
\end{figure}

\section{METHODOLOGY}
\label{sec:method}
To infer a kernel $\mathcal{A}$ from the observational data, a pair consisting of an observation $\mathcal{Y}$ and its corresponding activation map $\mathcal{M}$, we propose a two-step training strategy which results in encoder-decoder for inference shown in Fig.~\ref{fig:framework}.
In \textbf{Step 1}, a kernel VAE is trained to learn informative features from the kernels and accurately reconstruct them. Once training in Step 1 is complete, we freeze the kernel encoder and decoder. 
In \textbf{Step 2}, an \obsencoder is trained to extract meaningful latent representations of the corresponding kernel, 
which should be aligned with the latent representations produced by the well-trained \kencoder from Step 1 (i.e., $\textbf{h}_y$ and $\textbf{h}_A$ should be similar). During \textbf{inference},  a new observational data pair $(\mathcal{Y}, \mathcal{M})$ is first 
processed by the \obsencoder trained in Step 2. The resulting latent representations  then pass through the \kdecoder, which comes from Step 1, to produce the final output, an image of the inferred kernel.

\subsection{Kernel Encoder-Decoder}

To learn effective latent representations for kernels, we train a
\kencoder mapping each kernel $\mathcal{A}$ to a latent vector $\mathbf{h}_A$, which captures the underlying structure of $\mathcal{A}$. Subsequently, the \kdecoder reconstructs a kernel $\hat{\mathcal{A}}$ from $\mathbf{h}_A$:
\begin{equation}
    \mathbf{h}_A =  \textrm{Encoder}_{K} (\mathcal{A}); \;  \hat{\mathcal{A}} =   \textrm{Decoder}_{K} (\mathbf{h}_A),
\end{equation}
where the $\textrm{Encoder}_{K}$ and $\textrm{Decoder}_{K}$ represent the \kencoder and the \kdecoder respectively.
$\textrm{Encoder}_{K}$ and $\textrm{Decoder}_{K}$ are trained jointly by minimizing the difference between the reconstruction and the original kernel: 
\begin{equation}
\mathcal{L}_{\text{MSE}} = \frac{1}{H\times L}
(\hat{\mathcal{A}} - \mathcal{A})^2,
\end{equation}
where $H$ and $L$ represents the height and width of the image. Furthermore, taking account of the symmetrical characteristic of all kernels, the ideal reconstruction should also be well-symmetric\cite{tong2023scorebasedgenerativemodelsphotoacoustic}. Therefore, if the reconstruction is rotated by the angle between two axes of symmetry of the kernel, it should also be similar to the original one, for which a symmetric loss can be introduced: 
\begin{equation}
\mathcal{L}_{\text{SYM}} = \frac{1}{H_r\times L_r}
(\hat{\mathcal{A}}_{rotated} - \mathcal{A})^2,
\end{equation}
where $H_r$ and $L_r$ are the height and width of the selected window. We choose 40 for both in our experiments.

Overall, the training objective of Step 1 can be formalized as:
\begin{equation}
\label{eq:loss1}
\mathcal{L}_{\text{step1}} = \mathcal{L}_{\text{MSE}} + \mathcal{\alpha} \mathcal{L}_{\text{SYM}}+ \beta\mathcal{D}_{\text{KL}},
\end{equation}
where $\mathcal{D}_{\text{KL}}$ is the Kullback–Leibler divergence specific to variational autoencoders, $\alpha$ and $\beta$ are the weighting factors for different terms.
After training, this encoder-decoder is frozen, of which the $\textrm{Encoder}_{K}$  serves as a reference for computing latent representations of kernels in Step~2, and $\textrm{Decoder}_{K}$  is employed for inference.

\begin{figure}[t]
  \centering
  \includegraphics[width=0.8\linewidth]{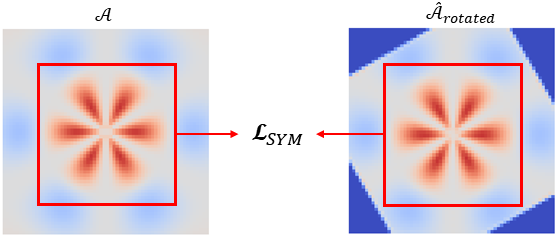} 
  \caption{Illustration of the symmetric loss $\mathcal{L}_{\text{SYM}}$:
           the reconstructed image is rotated by
           $180^\circ$, $120^\circ$, $90^\circ$, or $60^\circ$
           when there are $2$-, $3$-, $4$-, or $6$-fold symmetries, respectively. Only the values in the red box are taken into consideration.} 
  \label{fig:framework-sym}
    \vspace{-0.2in}
\end{figure}

\subsection{Observation and Activation Map Encoder}
In order to obtain the features for inferring the correct kernel from the observational data, another encoder is trained to align its output with the latent representation generated by $\textrm{Encoder}_{K}$.
Therefore in \textbf{Step~2}, we train the \obsencoder ($\textrm{Encoder}_{y}$) to map each observational data pair $(\mathcal{Y}, \mathcal{M})$  into the shared latent space of step1. To transform the observational data pair into a form acceptable to the encoder,   $\mathcal{Y}$ and 
$\mathcal{M}$ are stacked as a two-channel input, and fed into the trainable encoder:
\begin{equation}
    \mathbf{h}_y =  \textrm{Encoder}_{y} (\mathcal{Y},\mathcal{M}),
\end{equation}
where  $\mathbf{h}_y$ is the output latent vector, and $\textrm{Encoder}_{y}$ 
is implemented with the same structure as $\textrm{Encoder}_{K}$. 
To align the latent vector of observational data with that of the corresponding kernel, the $\textrm{Encoder}_{y}$ could be trained by minimizing the L2 loss\cite{_zdemir_2023}:
\begin{equation}
\mathcal{L}_{\text{step2}} =  ||\mathbf{h}_y - \mathbf{h}_A||_{L2}, 
\end{equation}
where $\mathbf{h}_A$ denotes the latent representations of the kernel $\mathbf{A}$ corresponding to the input observational data, which are obtained through the frozen $\textrm{Encoder}_{K}$ trained in Step 1. 
This objective encourages the model to embed the observed QPI patterns into a latent space that is consistent with the kernel embeddings, thereby facilitating the learning of meaningful mappings from observational data to their underlying scattering kernels.

    
\subsection{Inference Procedure}
As shown in Fig. \ref{fig:framework}, the whole model is set frozen at the inference stage. Given a new input pair $(\mathcal{Y}, \mathcal{M})$, $\textrm{Encoder}_y$ trained in Step~2 is used to compute the predicted latent representations of the kernel,
which are then passed to $\textrm{Decoder}_K$ (trained in Step~1), to finally generate the inferred kernel image. Notably, the kernel encoder $\textrm{Encoder}_K$ is not involved during inference.

\section{DATASETs}

\subsection{Dataset Overview}
In an experimental STM image, the QPI signal appears as oscillations centered around the scattering location. These signals often exhibit rotational symmetries, such as 2-, 3-, 4-, or 6-fold, with the exact symmetry and finer details determined by the material’s properties. The main challenge in isolating the QPI pattern of a single scatterer is that signals from many scatterers overlap and interfere across the material, making them difficult to disentangle in STM measurements.

To produce a dataset that  is both realistic and effective for training models on QPI kernel extraction, three requirements must be met: \underline{(1)} the dataset must be diverse and  abundant to cover a wide range of scattering conditions; \underline{(2)} each observation image must have an exactly known corresponding kernel to provide reliable supervision; and \underline{(3)} the data must realistically capture the QPI physics of interest, including material-specific features and experimental noise.
To this end, we design a dataset of 100 unique kernels with 500 observations each for a total of 50,000 samples. Every sample has a real-space image of the electronic structure (observation), a real-space image of the impurity configuration (activation map), and a real-space image of the corresponding kernel. 


\subsection{Dataset Construction} 
This immediately rules out experimental STM data as training material, even though it clearly satisfies requirement (3). Requirement (1) fails because STM experiments are costly and slow, often taking months to produce even a thousand images, many of which lack sufficient quality. Requirement (2) fails because the abundance of scatterers in real samples makes direct kernel extraction impossible with a good signal-to-noise ratio. These limitations motivate us to construct a dataset by the use of computational methods. Prior studies have simulated QPI data using physical models under experimentally realistic conditions, providing a promising alternative for scalable dataset construction 
\cite{sharma2021multi,doi:10.1021/acs.nanolett.4c01315}. 

These models are typically constructed in two parts: one component generates the kernel, and the other simulates scattering  from multiple kernels using known Quantum Scattering Theory \cite{crommie1993imaging}. 
Because they are grounded in physics, such models naturally satisfy requirement (3), and the simulated data also satisfy requirement (2) since the true kernel is known. However, these models are tightly bound to a specific physical system and therefore lack generality across diverse kernels, causing them to fall short of requirement (1). This limitation motivates us to  adopt  a phenomenological model that  reproduces the statistical behavior observed in experimental data while abstracting away from the specific underlying physics of kernel formation. Like the physical approach, our data construction model is also composed of two parts. The first component simulates a kernel according to the phenomenological assumptions:
\begin{equation}
\mathrm{Kernel}(r,\theta)
  = \sin^{2}\!\bigl(A\theta + \varphi_{1}\bigr)\,
    \sin^{2}\!\bigl(Br + \varphi_{2}\bigr)\,
    e^{-c r^{2}}
\label{eq:kernel}
\end{equation}
where $r$ is the distance from the defect location, and $\theta$ is the angle with respect to the defect location. The free parameters are: 1) $A$ selects the rotational symmetry ($0$ for total rotational symmetry, $\{1,\tfrac{3}{2},2,3\}$ for $(2,3,4,6)$-fold respectively); 2) $\varphi_{1}$ adjusts the orientation of the pattern ($\varphi_{1}$ is $\pi/2$ if $A=0$); 3) $B$ is the frequency of the QPI pattern;  4) $\varphi_{2}$ is the phase shift corresponding to the QPI oscillation; and 5) $C$ is related to the decay length of the QPI pattern.

\begin{figure}[t]
    \centering
    \begin{subfigure}{0.445\textwidth}
        \includegraphics[width=\linewidth]{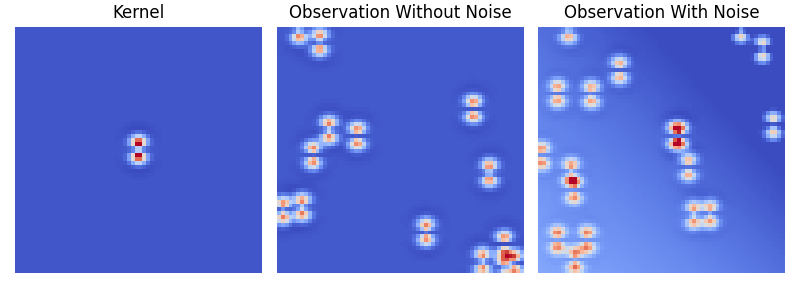} \vspace{-0.3in}
        \caption{2-fold} \vspace{+0.05in}
    \end{subfigure}
    \begin{subfigure}{0.445\textwidth}
        \includegraphics[width=\linewidth]{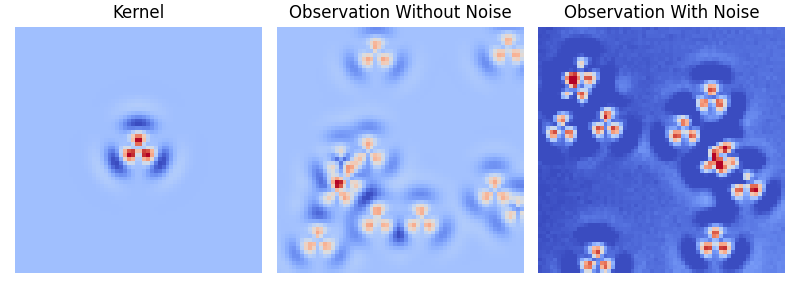}\vspace{-0.15in}
        \caption{3-fold}\vspace{+0.05in}
    \end{subfigure}
    
    \begin{subfigure}{0.445\textwidth}
        \includegraphics[width=\linewidth]{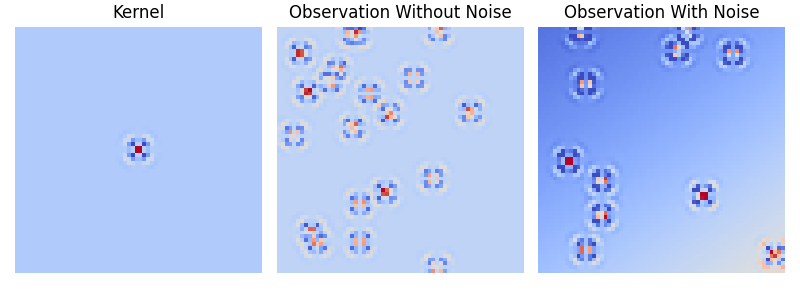}\vspace{-0.15in}
        \caption{4-fold}\vspace{+0.05in}
    \end{subfigure}
    \begin{subfigure}{0.445\textwidth}
        \includegraphics[width=\linewidth]{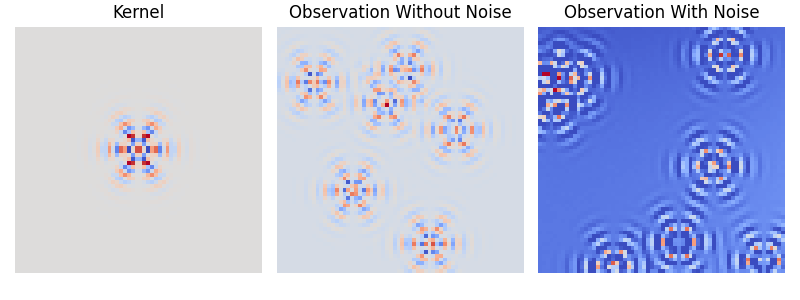}\vspace{-0.15in}
        \caption{6-fold}
    \end{subfigure}
    \caption{Comparison of noise-free and noisy datasets in a $20\times20\;\mathrm{nm}$ observation window, illustrating kernels with 2-, 3-, 4-, and 6-fold rotational symmetry.}
    \vspace{-0.2in}
    \label{fig:noisy_dataset}
\end{figure}

The second component of our model linearly superimposes multiple kernels to generate a complete observation. This simplified construction readily generalizes to arbitrary kernel shapes, thereby satisfying requirement (1). As with all simulated data, requirement (2) is automatically satisfied since the ground-truth kernels are known. Although the model is not strictly physical, it nonetheless fulfills requirement (3) by faithfully capturing the essential features observed in experimental QPI images. The key advantage is that this approach provides a realistic representation of QPI physics without being tied to the constraints of system-specific physical modeling.

The constructed dataset consists of 100 unique kernels, where $A$ is evenly distributed across each of the potential values, and $\varphi_{1}, B, \varphi_{2}, C$ are randomly selected within physically reasonable ranges. For each kernel, we generate 500 images. In each image, a random number of scattering centers (between 1 and 20) are placed at randomly chosen positions within a $40\times40\;\mathrm{nm}$ observation window with a resolution of $64\times64$ pixels. 
The pixel intensity is then computed by summing Eq.\ref{eq:kernel} over all scatterers, effectively superimposing the individual single-defect patterns into a complete QPI observation.
Fig.~\ref{fig:noisy_dataset} illustrating kernels with 2-, 3-, 4-, and 6-fold rotational symmetry, along with representative observations generated from each example kernel.

\begin{figure}[t]
  \centering

    \includegraphics[width=0.9\linewidth]{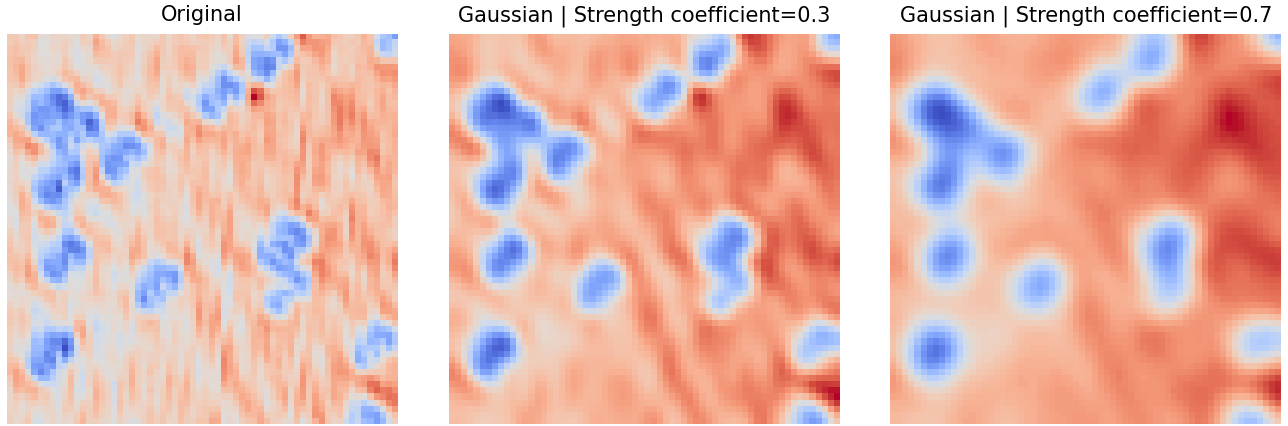} 
    \includegraphics[width=0.9\linewidth]{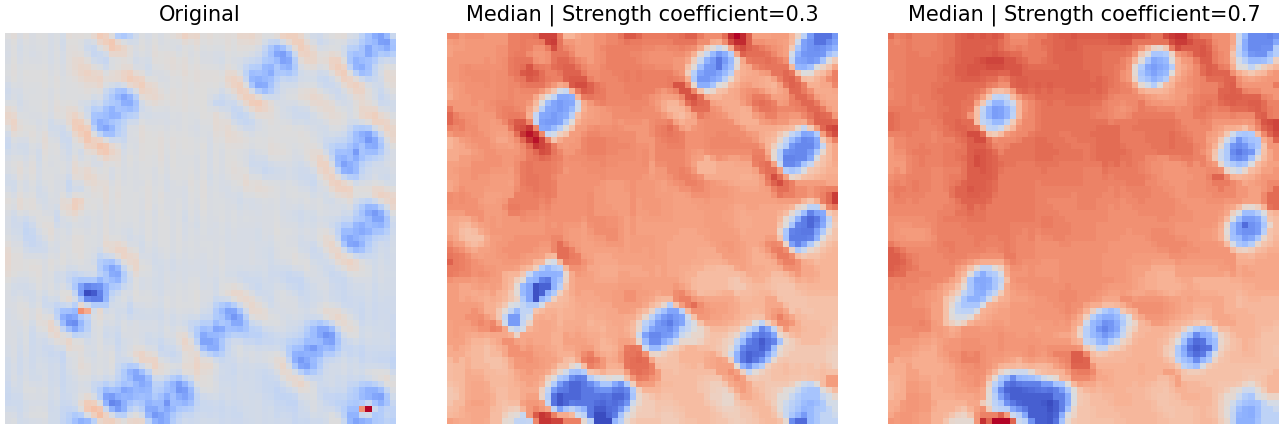}
    \includegraphics[width=0.9\linewidth]{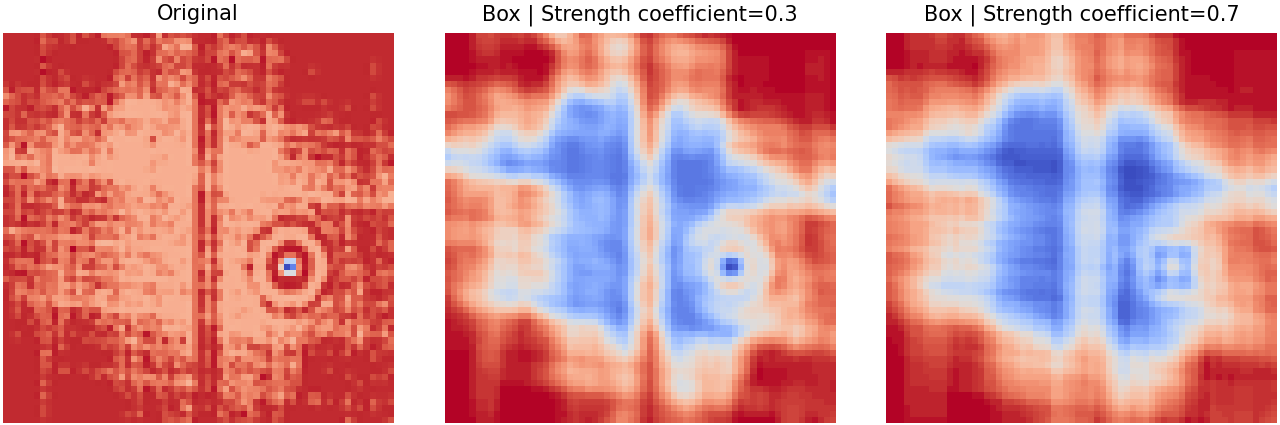}

  \caption{Examples of experimental observation denoised with different methods (1st row-Gaussian, 2nd row-Median, 3rd row-Box) at different intensities (1st column-original, 2nd column-light, 3rd column-heavy)}
  \label{fig:denoise_visualization}
    \vspace{-0.2in}
  
\end{figure}

To minimize the gap between simulated data and real-world experiments, we augment the synthetic dataset with carefully designed noise processes. In practice, experimental QPI images are inevitably affected by significant and highly non-uniform noise introduced by STM instrumentation. Factors such as thermal drift, mechanical vibration, electronic background, and detector limitations generate complex distortions that vary across both spatial and frequency domains, often overlapping with the true QPI signal. These irregular noise characteristics cannot be effectively removed by conventional denoising algorithms (see Fig.~\ref{fig:denoise_visualization}), which typically assume simpler or more homogeneous noise distributions. To better approximate experimental conditions, we augment our simulated images by adding polynomial backgrounds (smooth, low-frequency intensity variations unrelated to the QPI pattern), and injecting Gaussian noise. This design allows the synthetic dataset to mimic the intricate imperfections present in real STM measurements. As a result, models trained on this dataset are encouraged to generalize more effectively from controlled simulations to noisy experimental data.

The last column of 
Fig.~\ref{fig:noisy_dataset} shows the noisy observations, in comparison with those noise-free observations.  As illustrated, these perturbations substantially alter the clarity of the QPI patterns while preserving their underlying structure, making the task of kernel extraction more challenging and realistic. 

\subsection{Real Data Collection}
To extend the evaluation beyond the simulated data, we also collect real-world QPI measurements to test the validity of our proposed model. The data consists of QPI images taken on a low temperature STM on FeSe and Ag-111 samples at various sample bias voltages. Uncontrolled physical contributions such as thermal drift and electronic background manifest as combinations of both polynomial backgrounds and gaussian noise. This noise often overlaps with QPI oscillations, making the pattern nonuniform. For atomically resolved STM, images also contain the atomic lattice. In addition, the STM dataset has a different distribution from what was used to train the model, meaning the size of the observation window and pixel resolution is different between the experimental data and the simulated data. 

With the above considerations, we processed the experimental data in the following ways: we used denoise algorithms to remove the atomic lattice, cropped the STM images to $40\times40\;\mathrm{nm}$ observation window and resampled to a resolution of $64\times64$ pixels.
These modifications to the experimental data enables us to validate the degree to which our phenomenological dataset reflects the complexity and diversity of real measurements. The Ag-111 images offer relatively clean and well-understood oscillatory patterns, while the FeSe data captures more intricate electronic environments typical of quantum materials. Together, these datasets of 81 images of FeSe and Ag serve as a crucial reference point for assessing the realism and transferability of our simulated dataset.

\section{EVALUATION EXPERIMENTS}
\subsection{Experiment Setup}

\textbf{Training/Testing Splitting.} We adopt a structured data splitting strategy based on an 80/20 ratio for training and testing. First, the original kernel dataset comprising individual kernel images is divided into a kernel training set (80\%) and a kernel testing set (20\%). 
For the observation data ($\mathcal{Y}$ and $\mathcal{M}$), we construct two distinct test sets to evaluate the proposed method under both \textit{In-Domain} and \textit{Out-of-Domain} (OOD) generalization settings:

\begin{itemize}
    \item \textbf{In-Domain (ID) Setting:} We take  $\mathcal{Y}$ and $\mathcal{M}$   
    corresponding to kernels in the kernel training set and further split them into observed-data training and testing sets. This setting evaluates the model's ability to infer seen kernels from unseen observation with unknown scattering centers. 
    
    \item \textbf{Out-of-Domain (OOD) Setting:} We use  $\mathcal{Y}$ and $\mathcal{M}$  associated with kernels in the kernel testing set to construct the OOD test set. This setup tests the model’s ability to generalize to entirely new kernels and their corresponding QPI patterns, which were not seen during training.
\end{itemize}
\vspace{+0.05in}
Using the earlier ripple analogy, the  ID  setting corresponds to inferring \emph{which known stone caused new ripple patterns from different unknown positions}.  In contrast,  OOD  setting involves \emph{inferring unseen types of stones}, each producing a different fundamental ripple.

\textbf{Evaluation Metrics and Baselines.} To quantitatively assess the quality of the inferred kernel images, we employ three widely-used regression metrics: Mean Absolute Error (MAE), Mean Squared Error (MSE), and Root Mean Squared Error (RMSE), measuring the deviation between the predicted and ground truth kernels. 


Since our approach is the first machine learning solution for QPI kernel extraction, we compare it against the one-step baseline approach. Thus the evaluated approaches are:

\begin{enumerate}
    \item \textbf{One-Step Baseline:} A single VAE is trained to directly map   $\mathcal{Y}$ and  $\mathcal{M}$ to $\mathcal{A}$ (as shown in Fig. \ref{fig:example}). 
    
    \item \textbf{Our Two-Step solution}: as introduced in Sec. \ref{sec:method}. 
\end{enumerate}

\textbf{Implementation Details.} 
For both the one-step baseline and our two-step approach, we employ a pre-trained VAE (\texttt{madebyollin/sdxl-vae-fp16-fix})~\cite{von-platen-etal-2022-diffusers} from \textit{Hugging Face}, implemented with the \texttt{AutoencoderKL} architecture. 
The model adopts a U-Net style structure with 31 trainable layers (linear and convolutional) in the encoder and 51 layers in the decoder. 
The latent space has a dimensionality of $4 \times 8 \times 8$, and SiLU is used as the activation function throughout. 

The pre-trained VAE is fine-tuned in Step~1 to adapt to QPI kernels, while in Step~2 another observation encoder with the same architecture as the encoder of the former VAE is trained. 
The input layer is adjusted to accept either one-channel inputs (for kernel-only images in Step~1) or two-channel inputs (observation--activation pairs $(Y, M)$ in Step~2). 
All images are $64\times64$ pixels and normalized to the range $[-1,1]$ before being fed into the model. 

Training is conducted using the Adam optimizer (hyperparameters by default) with a learning rate of $1\times10^{-4}$, and batch size 8. 
Step~1 and Step~2 are trained for up to 300 and 50 epochs, respectively. 
The coefficients $\alpha$ and $\beta$ in Eq.~(4) are empirically set to 0.7 and $1\times10^{-6}$ to balance symmetry preservation and KL regularization. 

We adopt the pre-trained \texttt{sdxl-vae} due to its robust latent representation quality and convergence stability, which facilitate efficient transfer to the QPI domain without requiring large-scale retraining.


\subsection{Evaluation Results on Simulated Data} 
We evaluate both approaches under ID and OOD generalization settings. The results are reported in   Table \ref{tab:main}.

(1) In the \textbf{In-Domain} setting, both the one-step baseline and the proposed two-step method are capable of inferring kernel structures, reflecting the strong feature learning capacity of the variational autoencoder. However, the two-step method achieves significantly better inference quality. Specifically, it reduces the MAE from 0.0198 to 0.0118, MSE from 0.0033 to 0.0013, and RMSE from 0.0249 to 0.0211.

\begin{figure}[t!]
  \centering
        \includegraphics[width=0.9\linewidth]{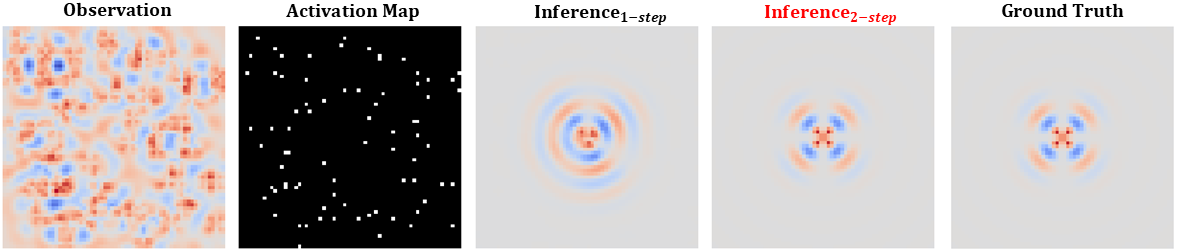}
                \includegraphics[width=0.9\linewidth]{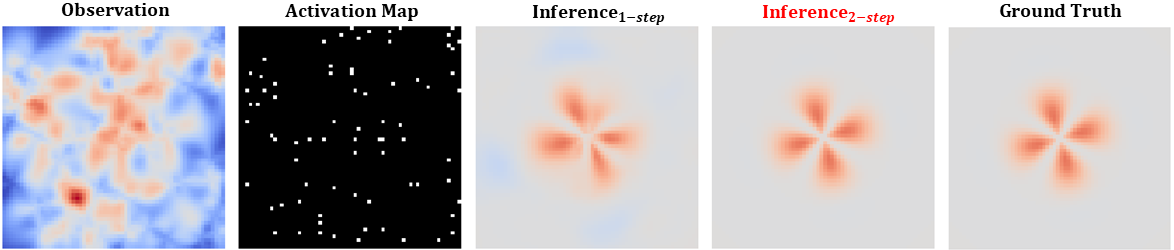} 

        \caption{Case study  on simulated samples. Comparison between baseline $\mathcal{A}_{1-step}$ and proposed $\mathcal{A}_{2-step}$, along with the ground-truth kernel $\mathcal{A}$. }
        

  \label{fig:case-study-comparison}
    \vspace{-0.25in}
\end{figure}

(2) In the \textbf{Out-of-Domain} setting, the one-step method struggles to generalize, as it relies heavily on patterns learned from direct mappings between observation data and known kernels in the training set. Its performance deteriorates markedly, with MAE rising to 0.1240, MSE to 0.0398, and RMSE to 0.1528. Qualitative inspection confirms that many predicted kernels significantly deviate from the correct mode. In contrast, the two-step method maintains stronger generalization, reducing the OOD MAE to 0.1013, MSE to 0.0294, and RMSE to 0.1297—substantially outperforming the baseline. Case study examples shown in 
Fig.~\ref{fig:case-study-comparison} confirm that our two-step method successfully captures an OOD kernel's four-fold symmetry (see the extraction result of our method $\mathcal{A}_{2-step}$ and the ground truth $\mathcal{A}$), whereas the one-step baseline fails to infer this unseen structure ($\mathcal{A}_{1-step}$).

\begin{table}[t!]
\caption{Performance of our two-step approach and the one-step baseline on different settings. (best values in bold)}
\label{tab:main} 
\centering
\resizebox{0.48\textwidth}{!}{\begin{tabular}{c|c|l|l|l}
\toprule[1.5pt]
\midrule
 Settings& Methods & MAE $\downarrow$    & MSE$\downarrow$& RMSE$\downarrow$\\ \midrule
\multirow{2}{*}{\centering   ID}& One-step Baseline     & 0.0198 & 0.0033 & 0.0249 \\
 & Two-step Approach& \textbf{0.0118} & \textbf{0.0013} & \textbf{0.0211} \\ \midrule
\multirow{2}{*}{\centering OOD} & One-step Baseline& 0.1240 & 0.0398 & 0.1528 \\
 & Two-step Approach& \textbf{0.1013} & \textbf{0.0294} & \textbf{0.1297} \\ \midrule
\bottomrule[1.5pt]
\end{tabular}} 
\end{table}

\begin{figure}[t!] \vspace{+0.1in}
  \centering
  \includegraphics[width=0.9\linewidth]{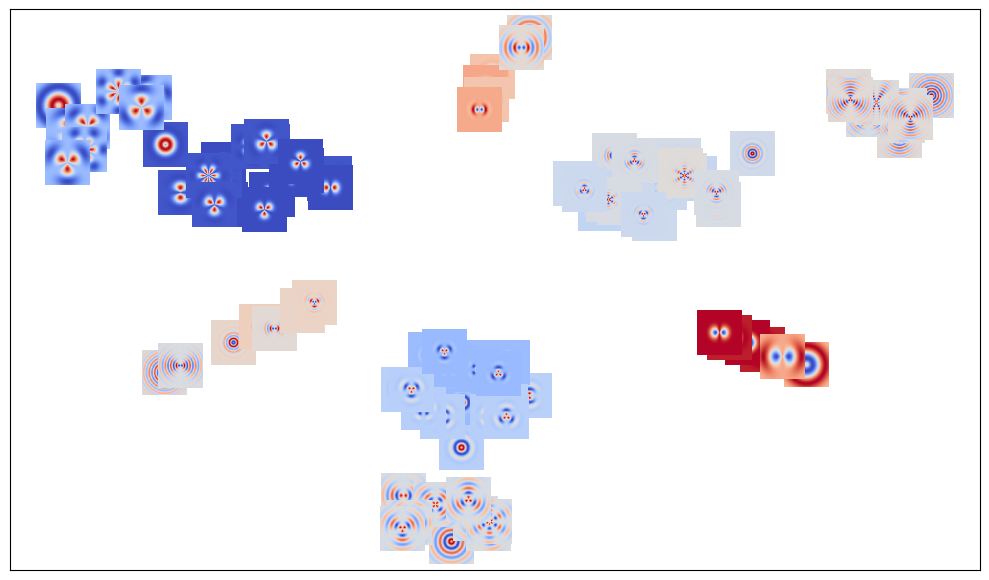} 
  \caption{t-SNE visualization of learned kernel representations, i.e.,  the vectors  $\mathbf{h}_A =  \textrm{Encoder}_{K}$ after training.}
  \label{fig:latent clusters}
    \vspace{-0.2in}
\end{figure}

We further examine the latent representations by applying t-SNE visualization on the vectors obtained by $\mathbf{h}_A =  \textrm{Encoder}_{K}$ after training (see Fig.~\ref{fig:latent clusters}). The resulting clusters consistently group together kernels that feature similar modes (i.e., simulation functions of similar amplitude), demonstrating that our \kencoder model ($\textrm{Encoder}_{K}$) inherently discriminates among kernels based on their underlying modes, and thus learns a representation that reflects meaningful physical structure.
This learned representation provides a structured latent space that guides the inference (decoding) process, enabling the model to recover kernels more accurately and robustly under complex multi-scatterer conditions.


\begin{figure}[t]
  \centering
        \begin{subfigure}[b]{0.4\textwidth}
        \includegraphics[width=\textwidth]{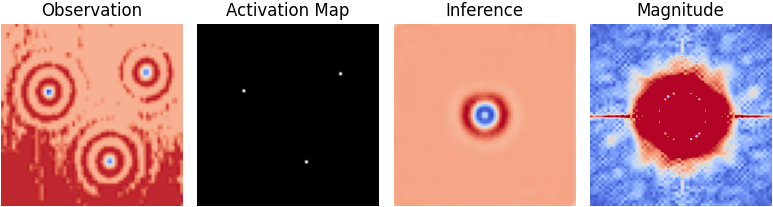 } 
        \includegraphics[width=\textwidth]{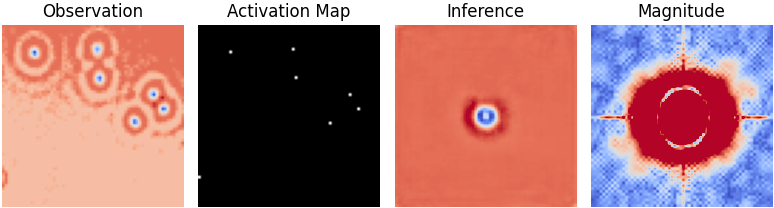}

        \caption{Ag 10mV}\vspace{+0.05in}
        \label{fig:exp_data ag}
        \end{subfigure}

        \begin{subfigure}[b]{0.4\textwidth}
        \includegraphics[width=\textwidth]{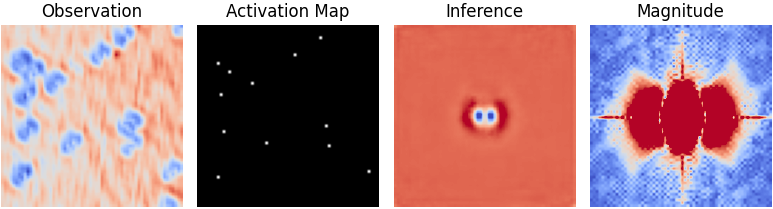 }
        \includegraphics[width=\textwidth]{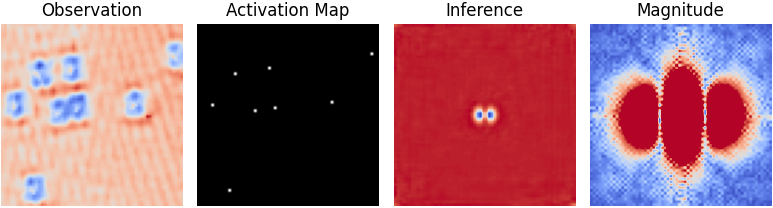}
          \caption{FeSe -10mV}
        \label{fig:exp_data fese}
         \end{subfigure} 
  \caption{Case study examples of real data.}
  \label{fig:experimental data}
    \vspace{-0.2in}
\end{figure}


\vspace{-0.1in}
\subsection{Evaluation Results on Real Data}
 To further test the effectiveness of the proposed method, we conducted experiments on real datasets (as shown in Fig.~\ref{fig:experimental data}). For real data observations, there are no ground truth kernels, however, the validity of our kernel extraction can be inferred from knowledge of QPI and previous reports. First, consider the symmetry of the Ag and FeSe kernels, which should exhibit total rotational (ring-like) symmetry and C2 (dumbell-like) symmetry respectively. The symmetry of the extracted kernels clearly follow what is expected as seen in Fig.~\ref{fig:exp_data ag} and Fig.~\ref{fig:exp_data fese}. Next, we can consider observations of the same material at the same energy. For this purpose, we take the magnitude Fourier transform (FT) for comparison. The FT of Ag at 10mV reveals that although the general shape is very similar, the kernel becomes elongated in one direction, thus losing rotational symmetry. The FT of FeSe kernels at -10mV show expected C2 symmetry, but they differ in intricate details. Considering the limited quality of real data, the kernel extraction extracted the kernel effectively, and we expect this to be a powerful method to find unknown QPI kernels. 
 
 For a direct comparison to  previous reports, we also display the magnitude Fourier transform of the extracted kernels. While the rotational symmetry of the Ag kernel compares favorably with previous reports, the extracted kernel is missing longer-range oscillations, which are even visible in the observation \cite{avouris1994real}. The extracted FeSe kernel also compares favorably in terms of general shape, but suffers from the same weakness as the Ag kernel extraction \cite{kostin2018imaging}, \cite{rhodes2019kz}, \cite{lin2023real}. The missing long-range oscillations lower the quality of the Fourier transform, which makes direct comparison to literature difficult, as many reports display the FeSe QPI signal in frequency-space. Physically, these patterns arise from electrons scattering off a nearby impurity and creating standing waves, so the symmetry and decay of the pattern is indicative of the material's electronic band structure. Considering that the model is trained on phenomenologically simulated data and limited data quality, extraction of the general shape of the QPI kernels from noisy STM data supports the robustness of this method. A future endeavor will involve training the model on data simulated by rigorous physical models.

\begin{table}[t]
\caption{Ablation study on symmetric loss} 
\label{tab:Ablation} 
\centering
\resizebox{0.46\textwidth}{!}{\begin{tabular}{c|c|l|l|l}
\toprule[1.5pt]
\midrule
  Settings& Methods & MAE $\downarrow$    & MSE$\downarrow$& RMSE$\downarrow$\\ \midrule
\multirow{2}{*}{\centering   OOD}& w/o sym loss& 0.1094 & 0.0329 & 0.1372 \\ 
& w/ sym loss     & \textbf{0.1013}& \textbf{0.0294}& \textbf{0.1297}\\
 \midrule
\bottomrule[1.5pt]
\end{tabular}} 
\end{table}

\begin{table}[t]
\caption{Ablation study on usefulness of pre-trained VAE} 
\label{tab:Ablation2} 
\centering
\resizebox{0.46\textwidth}{!}{\begin{tabular}{c|l|l|l}
\toprule[1.5pt]
\midrule
   Methods & MAE $\downarrow$    & MSE$\downarrow$& RMSE$\downarrow$\\ 
\midrule
 w/o pre-trained VAE& 0.1170 & 0.0556 & 0.1847 \\ 
 w/ pre-trained VAE& \textbf{0.0270}& \textbf{0.0029}& \textbf{0.0421}\\
 \midrule
\bottomrule[1.5pt]

\end{tabular}} 
\end{table}

\begin{figure}[t]
  \centering
        \begin{subfigure}[b]{0.45\textwidth}
        \includegraphics[width=\textwidth]{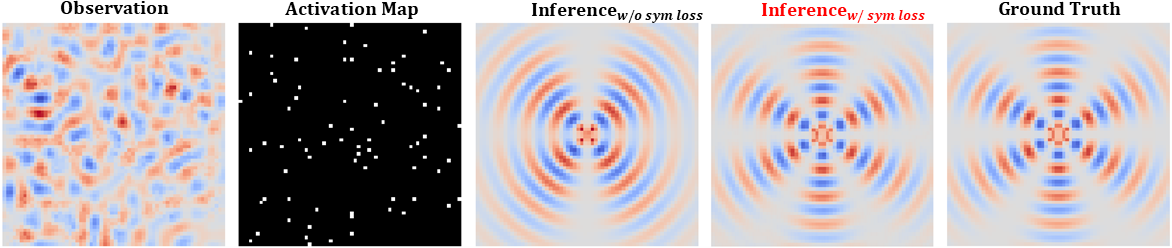} 
          \caption{Ablation on using $\mathcal{L}_{\text{SYM}}$} \vspace{+0.1in}
        \label{fig:ablation1}
        \end{subfigure}
        
        \begin{subfigure}[b]{0.45\textwidth} \centering
        \includegraphics[width=0.8\textwidth]{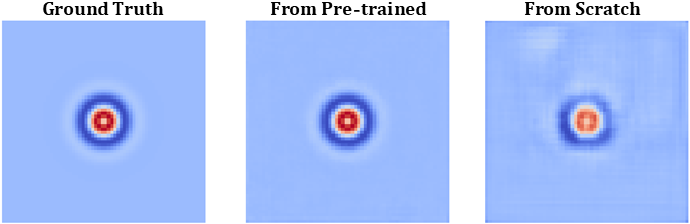}
        \caption{Ablation on pre-trained model.}
        \label{fig:ablation3}
        \end{subfigure}

  \caption{Examples for ablation studies.}
  \label{fig:ablation}
    \vspace{-0.2in}
\end{figure}

\vspace{-0.1in}
\subsection{Ablation Studies}
\textbf{Symmetric loss $\mathcal{L}_{\text{SYM}}$} (in~Eq. \ref{eq:loss1})  was proposed in training step~1, targeting a better capture of the kernel's symmetry. To assess this, we conduct an experiment to explore the performances between settings w/ and w/o $\mathcal{L}_{\text{SYM}}$,  as shown in Table \ref{tab:Ablation}. As we can see, there is a 6\% improvement on RMSE when symmetric loss is added, showing the effectiveness of the symmetric loss. The example in Fig.~\ref{fig:ablation1} also confirms the better inferred kernel with the help of the symmetric loss.

\textbf{Usefulness of Pre-trained VAE} was evaluated by comparing our model, which fine-tunes the pre-trained VAE,  with an alternative trained from scratch, the result in Step-1 for kernel reconstruction is shown in Table \ref{tab:Ablation2}.
The results in Fig.~\ref{fig:ablation3} indicate that 
  fine-tuning the pre-trained VAE leads to more accurate kernel extraction than training from scratch in the kernel reconstruction process.

\begin{figure}[t]
  \centering
  \begin{subfigure}[b]{0.48\textwidth}
    \centering
    \includegraphics[width=\textwidth]{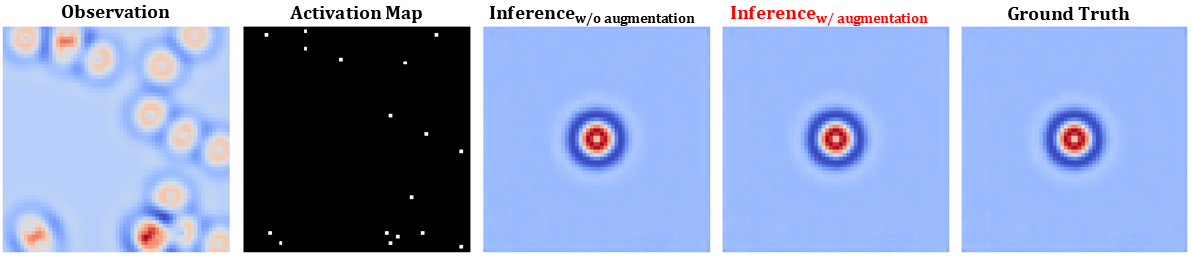}
  \caption{Noise-free data}
  \label{fig:Noise-free data}
  \end{subfigure}

  \vspace{0.1cm} %
  \begin{subfigure}[b]{0.48\textwidth}
    \centering
    \includegraphics[width=\textwidth]{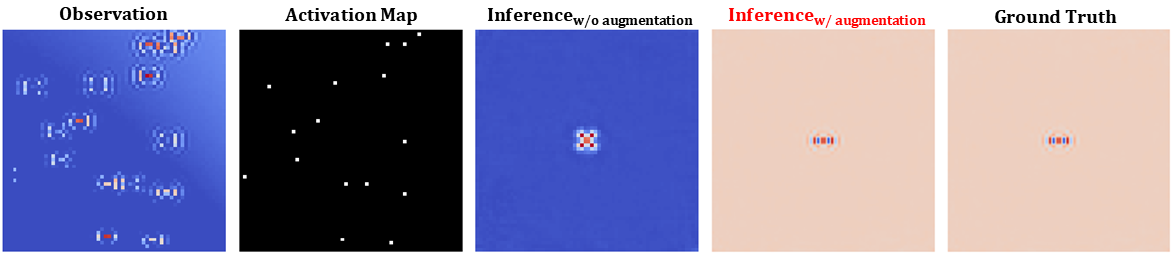}
  \caption{Noisy data}
  \label{fig:Noisy data}
  \end{subfigure}

  \vspace{0.1cm} %
  \begin{subfigure}[b]{0.45\textwidth}
    \centering
    \includegraphics[width=\textwidth]{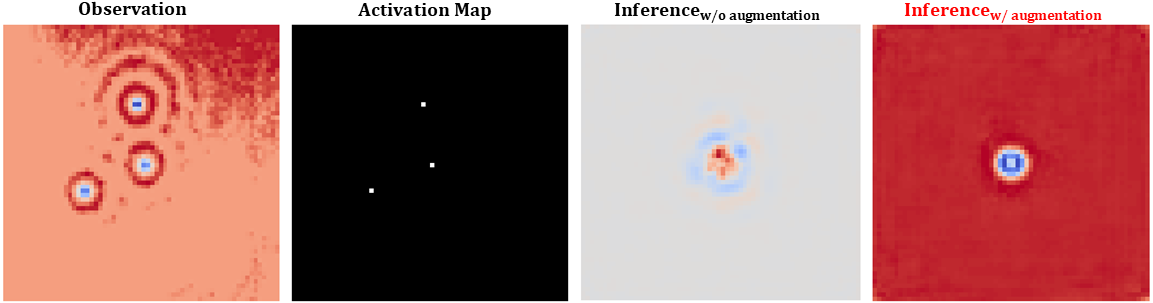}
  \caption{Real data}
  \label{fig:Real data}
  \end{subfigure}

  \caption{Ablation study on noise augmentation}
  \label{fig:noisy&noise_free}
    \vspace{-0.3in}

\end{figure}

\textbf{Noise augmentation} was introduced to bridge the gap between simulation and real data, enhancing both robustness and overall model performance. Figure~\ref{fig:noisy&noise_free} presents ablation studies comparing two models: one trained on the original dataset without noise and the other on the augmented dataset with noise. Both models were evaluated on noise-free simulated data, noisy simulated data, and real data.

On real data (Fig.~\ref{fig:Real data}), the model trained with noise augmentation successfully predicts a reasonable pattern, whereas the model trained on the noise-free dataset fails to recover a valid kernel, showing strong susceptibility to the real noise. Moreover, the noise-augmented model maintains good performance even on noise-free observations (Fig.~\ref{fig:Noise-free data}), while the noise-free model breaks down when tested on noisy observations (Fig.~\ref{fig:Noisy data}). These results demonstrate both the necessity and the robustness of training with augmented datasets for real-data scenarios.
 

 \section{CONCLUSION AND FUTURE WORK}\label{sec:future work}
This paper presents a two-step approach based on the VAE architecture to address the QPI kernel extraction problem. 
Specifically, we first optimize an encoder-decoder network using kernel data, and then introduce another encoder trained on observational data in the next step.
We further propose a symmetric loss utilizing the symmetrical characteristic of kernels. The results in various experimental settings demonstrate the superiority of our proposed approach in solving inverse tasks. 
In future work, we will focus on generalizing our two-step approach to other applications involving inverse tasks in complex physical systems, such as tomographic reconstruction and source separation, where direct mapping from observational data may also be unstable or non-unique.

Beyond its immediate success on QPI, our data-driven two-step framework offers a generally applicable strategy for disentangling latent structure in complex inverse problems. By learning compact, interpretable latent representations of kernels and aligning them with observations, this approach can be adapted to diverse domains where extracting key patterns from noisy, convoluted data is needed. The scalability of deep generative models means this approach can handle ever-larger datasets and growing experimental complexity. Crucially for quantum materials research, the framework enables more efficient and transparent analysis of large QPI datasets. Leveraging the pre-trained latent space, scientists can rapidly infer scattering kernels and reveal underlying electronic features, significantly accelerating the analysis of experimental data. Reducing both computational and interpretive bottlenecks, this approach paves the way for faster discovery of new quantum materials and phenomena, exemplifying how machine learning-driven methods can transform data-intensive science.

\vspace{-0.1in}
\section*{Acknowledgment}
This work is sponsored by the STIR program of University of Notre Dame. 
\vspace{-0.15in}

\bibliographystyle{IEEEtran}
\bibliography{references}

\end{document}